\documentclass[a4paper, 10pt, conference]{ieeeconf}      
\usepackage{FG2025}

\usepackage{subcaption}

\usepackage{graphicx}
\usepackage{soul}
\usepackage{booktabs}
\usepackage{multirow}
\usepackage{color}

\makeatletter
\let\NAT@parse\undefined
\makeatother
\usepackage[backref=page,colorlinks=true]{hyperref}
\usepackage{cite}

\let\paragraph\oldparagraph
\let\subparagraph\oldsubparagraph
\usepackage{titlesec}
\def\etal{\emph{et al}.}
\def\ie{\emph{i.e.,}}
\def\eg{\emph{e.g.,}}
\def\sota{\emph{state-of-the-art}}

\def\intdataset{\textsc{InterHand2.6M}}

\usepackage[accsupp]{axessibility}  

\usepackage{etoolbox}

\newbool{highlightText}
\newbool{removeTextBool}
\setbool{highlightText}{false}

\newbool{highlightToRich}
\setbool{highlightToRich}{false}

\setbool{removeTextBool}{true}

\newcommand{\mymovedparagraphs}[1]{\ifbool{highlightText}{\textcolor{blue}{#1}}{#1}}
\newcommand{\newstuff}[1]{\ifbool{highlightText}{\textcolor{ForestGreen}{#1}}{#1}}
\newcommand{\respondtoreviewer}[1]{\ifbool{highlightText}{\textcolor{red}{#1}}{#1}}
\newcommand{\removetext}[1]{\ifbool{removeTextBool}{}{\st{#1}}}
\newcommand{\rewritten}[1]{\ifbool{highlightText}{\textcolor{brown}{#1}}{#1}}

\newcommand{\richcheck}[1]{\ifbool{highlightToRich}{\textcolor{red}{#1}}{#1}}

\FGfinalcopy 

\IEEEoverridecommandlockouts                              
\def\FGPaperID{230} 

\title{\LARGE \bf
HandOcc: NeRF-based Hand Rendering with Occupancy Networks
}

\author{\parbox{16cm}{\centering
    {\large Maksym Ivashechkin, Oscar Mendez, Richard Bowden}\\
    {\normalsize
    CVSSP, University of Surrey, Guildford, United Kingdom}\\
    {\normalsize \texttt {\{m.ivashechkin, o.mendez, r.bowden\}@surrey.ac.uk}}}%
}

\usepackage{fancyhdr}
\begin{document}

\ifFGfinal
\thispagestyle{empty}
\pagestyle{empty}
\else
\author{Anonymous FG2025 submission\\ Paper ID \FGPaperID \\}
\pagestyle{plain}
\fi
\maketitle

\thispagestyle{plain}
\pagestyle{plain}




\begin{abstract}
We propose HandOcc, a novel framework for hand rendering based upon occupancy.
Popular rendering methods such as NeRF are often combined with parametric meshes to provide deformable hand models.
However, in doing so, such approaches present a trade-off between the fidelity of the mesh and the complexity and dimensionality of the parametric model. 
The simplicity of parametric mesh structures is appealing, but the underlying issue is that it binds methods to mesh initialization, making it unable to generalize to objects where a parametric model does not exist. It also means that estimation is tied to mesh resolution and the accuracy of mesh fitting.
This paper presents a pipeline for meshless 3D rendering, which we apply to the hands. By providing only a 3D skeleton, the desired appearance is extracted via a convolutional model. We do this by exploiting a NeRF renderer conditioned upon an occupancy-based representation. 
The approach uses the hand occupancy to resolve hand-to-hand interactions further improving results, allowing fast rendering, and excellent hand appearance transfer.
On the benchmark{~\intdataset} dataset, we achieved~{\sota} results.
\end{abstract}

\section{INTRODUCTION}
\label{sec:intro}
As one of the most expressive parts of the human body, the hands play a crucial role in communication, interaction, and manipulation tasks, which drives the necessity for accurate and versatile hand estimation.
Hand pose estimation, as well as hand synthesis and rendering, is important to many areas, including: human-computer interaction, avatar generation, sign language production, and augmented or virtual reality applications such as teleoperation or telepresence.

There has been a significant body of work devoted to 3D hand pose estimation over the years.
The most prominent works are often monocular, exploiting image convolution,~{\eg}~\cite{spurr2018cvpr, spurr2021self, Moon_2020_ECCV_InterHand2.6M, zhang_3d_hand_pose}.
However, leveraging advances in technology, particularly GPU acceleration, enables us to achieve volumetric hand reconstruction alongside 3D rendering for novel view synthesis.

Most existing methods employ sparse 3D skeletal hand estimation, and for hand rendering they exploit mesh-based parametric representations such as MANO~\cite{MANO:SIGGRAPHASIA:2017}.
MANO parameterizes a 3D hand with a set of angles and shape coefficients that incorporate forward kinematics and generate a realistic hand mesh. 
It allows for the efficient estimation of the 3D volume of the hand, and the corresponding hand mesh can be used for rendering.
These properties, alongside a differentiable implementation, have made MANO a popular and widely used hand model.

Traditional hand-rendering methods often rely on texture maps and a colored mesh, where the hand geometry is controlled by a kinematic model, examples of such approaches were demonstrated in~\cite{Hasson2020LeveragingPC, Ge20193DHS, zhang2019end}.
Nevertheless, such methods have drawbacks.
For instance, relying on meshes can lead to mesh artifacts, limits on fidelity/detail, costly generation of personalized texture maps, and challenges in handling self-occlusions and intersections. 

Recently, the neural radiance field (NeRF)~\cite{mildenhall2020nerf} has gained a lot interest due to its ability  to represent the volume density and color space as a continuous function. 
Despite their recent popularity and speed, NeRFs are still attractive as they demonstrate excellent generalization to novel-view synthesis. This is due to their use of a continuous function. 
While the original proposal for NeRF was for static scenes, a lot of work has since explored extensions capable of integrating dynamics into the NeRF model.


\subsection{Related Work}
The articulation of the human body, especially hands, presents numerous challenges for neural rendering, particularly when generalizing across complex shapes and motions.
Some of the first approaches to adapting NeRF to a dynamic scene were D-NeRF~\cite{pumarola2020d} by Pumarola~{\etal} and Nerfies~\cite{park2021nerfies} by Park~{\etal}
Both methods are similar in the way they introduce a deformation field ({\eg} MLP network) that learns the transformation from a target scene to canonical space, and a canonical NeRF model that predicts colors and density.
\rewritten{The authors argue that a two-model approach that introduces a transformation from observation to canonical space is better than direct estimation and helps in generalization.
However, this technique also imposes additional constraints on models to learn information about the shared geometry between the canonical and observation space, and the corresponding appearance.}


To achieve better reconstruction and thus rendering for a human body, parametric models have gained popularity.
They allow integration of human geometry ({\eg} kinematics) into a neural model and can achieve more accurate and faster optimization.
We can divide them into two main types of human body parameterization: implicit or mesh-based representations.

The vital advantage of an \emph{implicit} parameterization is that it can be represented with a continuous and differentiable function.
Examples are the signed distance field (SDF) used in~\cite{Alldieck2021imGHUMIG, Karunratanakul2020GraspingFL, Park_2019_CVPR, Atzmon_2020_CVPR, wang2021neus}, occupancy maps~\cite{OccupancyNetworks}, implicit surfaces, point clouds, and transformation fields~\cite{PTF:CVPR:2021}.
For faster convergence and better accuracy, the implicit models are normally conditioned with input data such as sparse points ({\eg} skeletons) or mesh parameters.
The imGHUM approach~\cite{Alldieck2021imGHUMIG} of Alldieck~{\etal} utilizes skeleton points from the human body to condition the SDF.
Similarly, Karunratanakul~{\etal} in HALO~\cite{karunratanakul2021halo} exploit the hand skeleton to condition an occupancy network.
The NASA model~\cite{deng2019neural} by Deng~{\etal} presents a neural pose-conditioned occupancy approach based on a mesh, where high-quality surface details are learned using per-bone deformable transformation.
The PTF~\cite{PTF:CVPR:2021} of Wang~{\etal} extends NASA's approach to learning occupancy functions in the continuous rest-pose space by exploiting piece-wise transformation fields. 
Although NASA uses SMPL~\cite{SMPL:2015} mesh parameters as initialization, the PTF tries to robustly fit a mesh to the predicted point cloud.

\emph{Mesh} representations for the human body are extremely popular due to their differentiability, simplicity, and compactness.
They have been proposed for body (SMPL), hands (MANO~\cite{MANO:SIGGRAPHASIA:2017}), faces (FLAME~\cite{FLAME:SiggraphAsia2017}), and even animals~\cite{Zuffi20163DMM}.
One of the main advantages of a mesh-based model is that it provides a volumetric shape that exhibits aspects of realism.
The SMPL model is often utilized in body capture from images ({\eg}~\cite{Pavlakos2019ExpressiveBC, li2020hybrik}), and MANO in various hand estimation problems ({\eg}~\cite{zhe_hand, Zimmermann2019FreiHANDAD}).
However, the main problems with meshes are their coarse structure and low resolution, which are especially evident when it comes to mesh rendering.
One way to mitigate these issues involves increasing the number of faces to provide high-fidelity meshes~\cite{Luan_2023_CVPR, bib:handavatar}, at the cost of increasing computational complexity.
PHRIT~\cite{Huang_2023_ICCV} combines the advantages of a MANO mesh and implicit SDF to obtain a high-fidelity reconstruction at infinite resolution but lacks real-time inference.
Other techniques addressed the UV texture map of the mesh to improve hand color/texture~\cite{Chen2021I2UVHandNetIP}, or exploit graph convolution neural networks to obtain richer information about the hand surface~\cite{Ge20193DHS}.


Implicit models tend to take longer to converge, and most of the current implementations try to combine both (implicit and mesh) representations to exploit their advantages.
Moreover, with the continuous property of the NeRF that enables learning of density and color, hand-rendering methods provide accurate results with real-time efficiency. 
One of the first approaches to integrate NeRF for 3D hand rendering was LISA~\cite{corona2022lisa} of Corona~{\etal}
LISA exploited MANO parameters together with local bone coordinates to predict per-bone signed distance and color.
The signed distances contribute to the final volume densities that allow rendering.
However, due to the complexity of the mesh, the method struggles to perform in real-time.

HandNerf~\cite{Guo_2023_CVPR} by Guo~{\etal} presents a framework for 3D hand rendering utilizing NeRF.
It uses a deformation field to transform the input scene into a canonical space.
Additionally, HandNerf exploits a MANO mesh to query the closest facet for a 3D point, which is used to predict texture colors.
Similarly, LiveHand~\cite{mundra2023livehand} by Mudra~{\etal} uses mesh textures along with the distance to the mesh surface as an input to a NeRF to estimate hand density and color.
Both LiveHand and HandNerf exploit rendered MANO mesh depth to provide an extra loss and awareness to the NeRF model, along with the ray bounds determined by a 3D hand mesh volume.
The Hand Avatar~\cite{bib:handavatar} of Chen~{\etal} provides a high-resolution MANO-HD mesh with more faces and vertices.
Their method proposes a shading field, where anchors are used on the mesh to extract albedo information of the hand poses.
\removetext{Additionally, the authors exploit an occupancy map to minimize the self-occlusions of a single hand.}
Similarly, HARP~\cite{karunratanakul2023harp} explicitly model a parametric mesh-based hand with a normal map and albedo to tackle lightning conditions and articulation.

The hand appearance was mainly tackled by an implicit function that learns hand texture from multi-view images~\cite{corona2022lisa, Guo_2023_CVPR}. LiveHand embeds hand textures on the MANO UV texture map, while HandAvatar and HARP exploit the albedo and normal map from a MANO mesh.
Handy~\cite{Potamias_2023_CVPR} utilizes a GAN model to generate high-fidelity UV hand mesh textures.
In contrast, our work employs a separate Convolutional Variational Autoencoder (CVAE) to explicitly extract latent hand texture features directly from the desired image. These extracted features are then used to condition the rendering model, allowing for more control and flexibility.



\subsection{Motivation and Novelties}

\rewritten{We propose a framework for novel pose and novel view high-fidelity hand rendering.
Despite recent advancements in NeRF and rendering via Gaussian splatting, hand rendering remains an unsolved problem due to numerous challenges ({\eg} high motion, finger interactions, etc.). 
The current~{\sota} approaches rely heavily on the MANO mesh model, which introduces significant limitations: if the mesh is poor, the rendering quality suffers as well.
Additionally, obtaining accurate hand meshes is inherently challenging and often imprecise, requiring rendering and fitting to multi-view data.
These meshes are restricted by low resolution and coarse surface detail, which can lead to rendering artifacts.}


\rewritten{To avoid reliance on a parametric mesh model like MANO, we instead leverage an implicit shape representation by probabilistically modeling the occupancy of the hand.
This paper also serves as a proof of concept, demonstrating that an implicit model without mesh information achieves~{\sota} performance.
To the best of our knowledge, we are the first to render a dynamic hand using a pose-conditioned NeRF without relying on an underlying MANO model.
The main input to our model is as simple as a sparse 3D hand skeleton, which is easier to obtain than a volumetric mesh (using a basic triangulation).}

\rewritten{Furthermore, we remove the fundamental necessity for a parametric structure, which limits the application of current approaches to objects or body parts for which such models are unavailable.
Many existing methods rely heavily on explicit mesh textures, thus restricting their applicability to areas where models exist. By adopting an implicit shape representation, we avoid these constraints and open oportunties for more flexible, extendable, and accurate modeling.}

\begin{figure*}[t]
    \centering
    \includegraphics[width=0.99\linewidth]{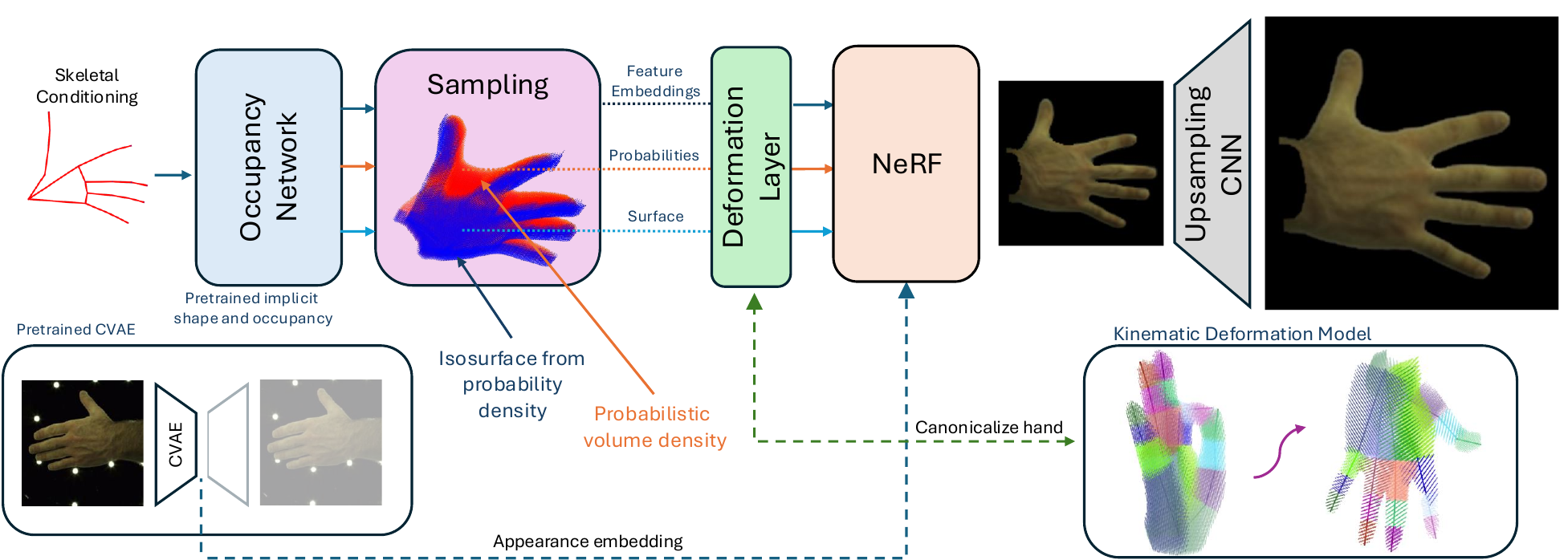}
    \caption{This figure demonstrates an overview of the proposed approach. Samples are drawn from the occupancy network which is conditioned on the skeletal input. The occupancy model returns per-point probabilities and features of the surface points. The surface points go through a deformation layer that canonicalizes the hand input. Afterwards, per-point occupancy encodings are then given to the NeRF (MLP) along with appearance embeddings. The NeRF renders an RGB image with corresponding features. The final CNN layer upsamples and refines the NeRF output.}
    \label{fig:pipeline}
\end{figure*}

We evaluate our model on the benchmark~{\intdataset} dataset~\cite{Moon_2020_ECCV_InterHand2.6M}, achieving {\sota} results. Additionally, by bypassing meshes, we propose a number of novelties and advantages that include: 
\begin{enumerate}
    \item Utilizing an occupancy map to facilitate efficient hierarchical sampling of a hand surface for NeRF rays.
    \item Conditioning the NeRF with CNN embeddings to provide improved hand appearance and shape transfer.
    \item Tackling interacting hands by modeling occupancies that prevent hand intersection.
    \item Efficient and fast rendering, where the hand geometry implicit to occupancy map is learned via a NeRF on a downscaled image, while exploiting a CNN for accurate image upsampling.
\end{enumerate}

\section{METHODOLOGY}
The overview of the proposed pipeline is demonstrated in Fig.~\ref{fig:pipeline}.
The ultimate goal is an accurate 3D rendering of an articulated hand from a single view.
We achieve this using a pre-trained occupancy network, a sparse 3D skeleton, and a NeRF renderer.
Additionally, we exploit appearance embeddings extracted with an image model, and an upsampling CNN that improves NeRF output.   


In the following sections, we discuss each part of the pipeline in turn. First, we cover the point cloud extraction and representation of the hand occupancy model. Next, we present hand appearance and shape transfer using a CVAE model. After that, we outline volumetric hand rendering with NeRF and the efficient hierarchical sampling of the hand surface. We also discuss the importance of the deformation model and canonical representation. Finally, we describe an upsampling CNN model that increases the rendered image quality.

\subsection{Point Cloud Extraction}

The first step in our pipeline is to extract dense point clouds from multi-view hand data to train an occupancy network. We assume \( N \) camera views with known calibration parameters \( \mathcal{P} \). Given an approximate 3D hand bounding box, we generate a uniformly distributed point cloud and use projection matrices \( \mathcal{P} \) to find the corresponding 2D projections of the 3D points.

Assuming color consistency of the hand pixels among images, we filter the points to identify those that correspond to the hand. To support this assumption, we normalize images for luminance, brightness, color, contrast, etc. This normalization ensures minimal color discrepancy across authentic hand projections. We can test this by calculating the standard deviation over \( N \) views for each point, with low variance indicating true hand projections.

While this filters hand points, background points with consistent colors may remain. We re-project all 3D points back into the 2D images and check their location with 2D binary hand masks to further filter incorrect points. 

\subsection{Hand Occupancy Representation}
Given the dense point clouds for each hand, the next step is to train an occupancy network.
The aim is to learn an approximate hand shape by conditioning the occupancy network with a sparse hand skeleton.

The occupancy network provides an implicit hand representation, where for each 3D point, it returns the probability of whether a point belongs to the hand.
If $\mathbf{x} \in \mathbb{R}^3$ is an arbitrary 3D point, $\mathbf{S} \in \mathbb{R}^{n\times 3}$ a sparse hand skeleton of $n$ points, then occupancy network $\mathbf{O}$ can be viewed as a probability function such that $\mathbf{O} : \mathbb{R}^3 \times \mathbb{R}^{n\times 3} \rightarrow [0,1]$.

We canonicalize the occupancy model to align with the same 3D space centered at the origin. Let $\mathbf{s}_0$ be the first point in the matrix $\mathbf{S}$ and the root joint of a hand ({\eg} wrist). The normalized hand skeleton $\Tilde{\mathbf{S}} = \mathbf{S} - \mathbf{1}_n\mathbf{s}^T_0$ conditions the occupancy map. Note that $\Tilde{\mathbf{s}}_0$ is at the origin,~{\ie} $\Tilde{\mathbf{s}}_0 = 0$. Moreover, we represent a left hand as a flipped right hand by mirroring points along the $x$-axis.

It is crucial to efficiently exploit the occupancy network in the case of interacting hands, especially to determine the probability of intersection.
$\mathbf{S}_R$ and $\mathbf{S}_L$ denote sets of points for the right and left hand, respectively.
$\mathbf{P}$ is a matrix of $K$ 3D points spanning both hands, and $\mathbf{t} = \mathbf{s}_0^L - \mathbf{s}_0^R$ is the offset between the hands.
The equation to determine the occupancy probability for the right hand is $\mathbf{O}(\mathbf{P}_i - \mathbf{s}_0^R, \Tilde{\mathbf{S}}_R)$ for all indices $i \in \{1,\dots, K\}$ of matrix $\mathbf{P}$.
For the left hand, the input points $\mathbf{P}$ must not only be shifted but also flipped and translated by the hand offset to preserve the original relation of the left to right hands.
Let $f$ be a function that flips input points by multiplying the $x$-axis by -1.
Then the occupancy map for the left hand of the same 3D point set $\mathbf{P}$ is:
\begin{equation}
    \mathbf{O}(f(\mathbf{P}_i - \mathbf{s}_0^L), f(\Tilde{\mathbf{S}}_L))
\end{equation}

\rewritten{The probabilities can serve as labels and strong indicators of which specific hand the points belong to, potentially determining a possible hand-to-hand intersection.
Crucially, the occupancy probabilities help set the boundaries of the hand and allow for the extraction of hand surfaces, which will be discussed in the following sections.}

\subsection{Hand Appearance and Shape Transfer}
Although the occupancy network is trained on point clouds derived from multi-view data, in practice, only a single image is provided as input.
Therefore, the hand parameters such as shape and appearance (skin color, wrinkles, gender, hair, nails, etc.) have to be extracted from a single image.

Many approaches parameterize the shape of the hand with MANO which exploits principal component analysis (PCA) to model hand shape.
As is common in the literature~\cite{Zimmermann2019FreiHANDAD, Li2022intaghand, Kulon_2020_CVPR}, mesh parameters are estimated from input images by supervising network training using ground truth mesh parameters. 
However, since we assume no mesh parameterization in the proposed pipeline, we demonstrate a new, alternative approach that extracts both shape and appearance from the images.
Moreover, it is also suitable for hand shape and appearance transfer.

\begin{figure}[t]
    \centering
    \includegraphics[width=0.99\linewidth]{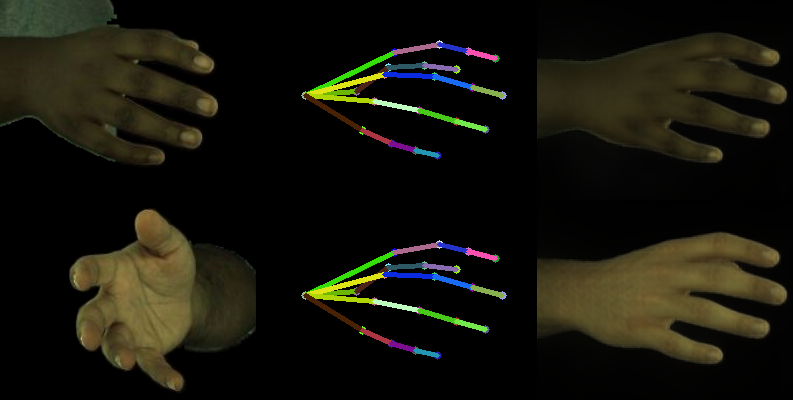}
    \caption{This figure demonstrates the results of the CVAE model. The first column shows input images $\mathbf{I}_{H_i}^A$ and $\mathbf{I}_{H_j}^B$ of two different hands from different people. The second column shows the same hand skeleton $\mathbf{H}_k^X$ rasterized on an RGB image. The output of the model is the last column, showing synthetically generated images $\mathbf{I}_{H_k}^A$, $\mathbf{I}_{H_k}^B$. These images closely resemble the input persons identity in terms of shared features, albeit with different skeleton shape.}
    \label{fig:cvae}
\end{figure}
First, let $\mathbf{H} \in \mathbb{R}^{n\times 2}$ be a 2D hand skeleton found by projecting the 3D hand skeleton $\mathbf{S}$, using known projection parameters $\mathcal{P}$ and $\mathbf{I}$ is an RGB image of a hand.
The objective is to learn a function $\phi$ capable of transforming a hand skeleton $\mathbf{H}$ and image $\mathbf{I}$ into a latent vector representing the hand composition.
This composition should capture both the skeletal structure of the hand $\mathbf{H}$ and its appearance in the image $\mathbf{I}$.

For this task, we exploit a convolutional variational autoencoder (CVAE)~\cite{kingma2022autoencoding}, because they are efficient generative models capable of accurately encoding the input into a latent representation that follows a Gaussian distribution. 
The CVAE aims to take the hand skeleton $\mathbf{H}$ and an image of a hand $\mathbf{I}$, then compress that information into a lower dimensionality latent space with sufficient detail that a decoder can reconstruct an image of a hand with the same appearance $\mathbf{I}$ and skeleton $\mathbf{H}$.

Let the function $\phi$ be the CVAE encoder, where the function $\psi$ decodes the latent vector of $\phi$ to an RGB image.
Then for any person $X$ and for any hand pair $(i, j)$:
\begin{equation}
\label{eq:main_rule}
    \phi({\mathbf{H}_i^X}, \mathbf{I}^X_{H_i}) = \phi(\mathbf{H}_i^X, \mathbf{I}^X_{H_j})\:\:\:\:\:\: \land \:\:\:\:\:\: \psi\big(\phi(\mathbf{H}_i^X, \mathbf{I}^X_{H_j})\big) = \mathbf{I}^X_{H_i}
\end{equation}
Where, $\mathbf{H}_i^X$ is $i$-th hand skeleton of the $X$-th person, and $\mathbf{I}_{H_j}^X$ is an image of the $\mathbf{H}_j$ hand of the $X$-th person. In other words, the latent space for the same hand skeleton and images of the same person are equal.
For two different identities $A$ and $B$, the goal of the CVAE is to be able to produce the following:
\begin{equation}
\label{eq:cvae_goal}
\psi\big(\phi(\mathbf{H}_i^A, \mathbf{I}_{H_j}^B)\big) = \mathbf{I}_{H_i}^B
\end{equation}
Which is a hand image generated of person $B$ with a hand skeleton of person $A$.
The CVAE is forced to disentangle the skeleton pose from the hand appearance of the image.
Consequently, by providing different people in the training set, the CVAE can generalize over various shapes and appearances.

Since each person has a different distribution of hand skeletons, in practice, Eq. (\ref{eq:cvae_goal}) cannot be used directly, because such images may not exist.
Therefore, the CVAE necessitates input from the same individual, as specified in Eq. (\ref{eq:main_rule}).
The associated losses for training serve to enforce consistency within a single person's latent space. They comprise the image loss of the CVAE decoder, and the Kullback-Leibler divergence~\cite{Joyce2011}.
Examples are demonstrated in Fig.~\ref{fig:cvae}.

\subsection{Hand Rendering}
The standard NeRF architecture processes a 5-dimensional input, consisting of 3D point coordinates, $\mathbf{x}$, paired with viewing direction, $\mathbf{d}$ (represented as a 3-dimensional vector).
It then estimates volume density $\sigma (\mathbf{x})$ alongside the RGB color vector $C(\mathbf{r})$. 
Leveraging multiple views, NeRF casts rays $\mathbf{r} = \mathbf{o} + t\mathbf{d}$  from pixel coordinates using camera parameters $\mathcal{P}$, where $\mathbf{o}$ denotes the camera origin.
Subsequently, it samples these rays within predefined bounds $t \in [t_{\min}, t_{\max}]$.
The predicted colors are accumulated from the colors and volume densities along the corresponding camera rays.


\begin{equation}
    C(\mathbf{r}) = \int_{t_{\min}}^{t_{\max}} T(t) \, \sigma\big(\mathbf{r}(t)\big) \, \mathbf{c}\big(\mathbf{r}(t), \mathbf{d}\big) \, dt
\end{equation}
\begin{equation}
 T(t) = \exp\bigg(-\int_{t_{\min}}^{t_{\max}} \sigma\big(\mathbf{r}(s)\big) ds\bigg)
\end{equation}

Where, the $T(t)$ function corresponds to the accumulated transmittance along the ray, and $\mathbf{c}$ is a color function of the ray and viewing direction. 
In the literature, the continuous NeRF function is implemented via an MLP.
In practice, the integral is approximated by weighing the discrete point samples along the ray.
To mitigate the dependency on the fixed ray bounds, hierarchical volume sampling is performed based on a coarse density estimate. 

We build our framework around the concept of NeRF's rendering approach, however, we adapt it to the articulated hand problem.
Hand motion is governed by a skeletal structure, while appearance and shape are determined by CVAE image embeddings $\mathcal{A} \in \mathbb{R}^n$.
\rewritten{Directly encoding a skeleton vector and 3D coordinates into a NeRF MLP is inefficient and results in poor conditioning, as the skeletal embeddings lack volumetric information, are sparse, and lead to poor generalization ({\eg} see also \cite{Guo_2023_CVPR}).}
Therefore, we leverage a pre-trained occupancy decoder features $\mathcal{F} \in \mathbb{R}^m$, where the input 3D points have already been processed via a skeleton occupancy embedding.
These features encapsulate spatial information and the relationship to the 3D skeleton, serving as input to the NeRF.

Furthermore, the occupancy, combined with per-point features, yields probabilities $\mathcal{P}$ that we employ as density cues to expedite NeRF convergence.
This is particularly beneficial for interacting hands, where probabilities facilitate the identification of specific hands by selecting the maximum probability.

Refining the NeRF output via a CNN model is one of our objectives, and several studies ({\eg}~\cite{mundra2023livehand,Guo_2023_CVPR}) enforced NeRF to produce additional $d$-dimensional features $\mathbf{f}_N \in \mathbb{R}^d$ alongside RGB colors and volume density.
This approach is adopted because simple RGB channels lack high-dimensional information regarding volumetric shape and details, which are crucial for the post-processing CNN model. Consequently, we incorporate this strategy into our framework as well.
Ultimately, NeRF can be conceptualized as a function $\boldsymbol{g}$ such that it takes appearance, occupancy features and probabilities, and returns color, density, and volumetric features:
\begin{equation}
    \boldsymbol{g}: (\mathcal{F}, \mathcal{A}, \mathcal{P}) \rightarrow (\mathbf{c}, \sigma, \mathbf{f}_N)
\end{equation}


In the original NeRF formulation, the authors randomly sample rays from images.
However, such a strategy restricts us to an RGB loss ({\eg} mean squared error, MSE) on the predicted and ground truth pixel colors.
To integrate a perceptual image loss that is superior in capturing detailed features ({\eg} LPIPS~\cite{zhang2018perceptual}) we use all image rays to fully render an image.
Nevertheless, loading all rays is very computationally expensive.
Therefore, for efficiency reasons, we prune rays with zero occupancy probability.
This allows us to perform batching over the images without a significant memory demand.

\subsection{Hand Bounds}

In the vanilla NeRF, the ray sampling bounds are set by user-defined nearest and farthest distances that span the object.
The ``coarse'' NeRF model uses uniformly sampled points along rays, and a ``fine'' model then provides hierarchical sampling.

For dynamic hands, using fixed bounds is inefficient because they must be large enough to span over all hand articulation, and to span the whole object's volume, many samples are required.
Therefore, we exploit the probabilistic occupancy network to establish constraints and sample points in close proximity to the hand surface.
We define the origin of the ray intersecting the hand surface based on an occupancy probability threshold $p_{\min}$ (e.g., 0.1). The upper bound of the ray can be determined by the point of maximum occupancy saturation (\emph{i.e.}, $p_{\max}$, close to probability 1.0), a predefined minimum distance (e.g., 1-2 cm), or, as implemented in our approach, a combination of both.
The estimated hand surface is essentially the closest point on a ray to the camera that has an occupancy probability equal to a pre-defined threshold.
Figure~\ref{fig:pipeline} depicts the hand surface represented by blue points.

\subsection{Occupancy Hierarchical Sampling}



\rewritten{The probabilities returned by the occupancy network can be interpreted as densities in a NeRF model. However, these occupancy probabilities are not reliable enough to directly replace the NeRF density field, as doing so would lead to degraded rendering quality. Instead, we leverage the occupancy field to incorporate additional sample points from regions with the highest probabilities by employing hierarchical sampling~\cite{mildenhall2020nerf}.}
This procedure mirrors the original proposal by the NeRF authors, which involved a fusion of ``coarse'' and ``fine'' models.
However, by sampling directly from the hand surface and augmenting it with points of the highest probabilities, we circumvent the need for a ``fine'' model.

\subsection{Deformation Model}
To improve generalization and accelerate training, we integrated a deformation model that transforms an observed hand pose into a canonical one.
This assists NeRF in optimizing a single canonical pose, rather than various hand poses, thereby reducing complexity.
We base our deformation only on skeletons.
Firstly, we determine the per-bone rotations and translations between the canonical and observed sparse skeletons.
Then, we segment a hand point cloud based on proximity to the underlying skeleton edges and transform the points using corresponding per-bone rotations and translations. Unlike other methods that utilize MLP-based deformation for enhancement or prediction, our approach does not require a deformation neural network. The per-bone rigid transformations are precise, and any artifacts resulting from the segmentation are either negligible or mitigated by filtering the point cloud through the occupancy network.

\subsection{CNN Upsampling}

As is often mentioned, NeRF can be slow to train and takes a long time to fully converge.
Even with the proposed efficient sampling, the time to render a full-resolution image takes significant computational time.
Since NeRFs are capable of accurately representing volumetric geometry and high rendering quality, these properties can be exploited at a much lower image resolution.
Moreover, upscaling CNN models have shown excellent capability in restoring and refining down-scaled images.
Therefore, we enhance rendering speed by training a NeRF on low-resolution hand geometry and subsequently reconstructing the full-size image using a CNN.
This process involves providing a provisional RGB image along with additional feature channels $\mathbf{f}_N$.


\section{EXPERIMENTS}
We primarily focus our evaluation on the publicly available {\intdataset} benchmark dataset.
It is one of the largest hand datasets containing 2.6 million images captured with 140 cameras of 26 unique people.
The dataset is commonly used to evaluate 3D hand estimation and rendering.

\subsection{Implementation Details and Reproducibility}
To extract hand point clouds from multi-view images, we combined hand masks and color consistency to filter out outliers.
The images are normalized with histogram equalization and converted to the HSV format.
Since ground truth point clouds are unavailable, visual verification on a random subset of hands confirms their high realism.
For our occupancy network, we leverage a PointNet encoder~\cite{pointnet} to condition it with the sparse 3D input skeleton data. 
The decoder utilizes conditional batch normalization with ResNet~\cite{resnet} blocks to convolve the input 3D points, incorporating the embedded skeleton from the encoder, and transforming the points into logits. 
The intersection over union accuracy of our occupancy network is 80\% on the validation set.
Such performance demonstrates high overlap with the ground truth hand occupancy, enabling reliable shape reconstruction.



For the CVAE model, we used a CNN with residual blocks to downsample and upsample the images.
The latent space estimated by the encoder is further reduced by PCA. The loss for the CVAE combines LPIPS, L1 and KL Divergence. 
%
%
For the NeRF, we use 8 uniformly distributed point samples along the ray and another 8 points from hierarchical occupancy sampling.
The NeRF has a width of 256-512 hidden units for the MLP, with a depth of 8 layers. 
The CNN upscaling model is SRResNet~\cite{Ledig_photo_2017} and has a scale factor of 2.
The model is trained end-to-end using a combined LPIPS (0.4) and L1 (0.6) loss on the images. 
All models were trained until convergence using an Adam optimizer~\cite{Kingma2014AdamAM} with an NVIDIA GeForce RTX 3090 GPU. At inference, our unoptimized model renders at 7fps in comparison to a simple mesh render at 58fps.  

\subsection{Affine Image Transform}
\hfill\\
Using the original downscaled image in NeRF is inefficient, as it often contains a large amount of empty background, with the hand occupying only a small portion of the space.
Using the approximate 2D bounds of the hand (found from a 2D skeleton), we crop the original image into a fixed bounding box via an affine transform $\mathbf{T} \in \mathbb{R}^{3\times 3}$.

\removetext{Since there are many ways to transform a hand image, we exploit this to provide additional augmentation for the NeRF.}
During training, we introduce a random scale or translational shift to the affine transform $\mathbf{T}$.
Such an augmentation casts new rays, providing subpixel precision.
It helps the NeRF to generalize and provides new observations rather than overfitting to the same rays every time.
To preserve the original image coordinates, we update the corresponding intrinsic matrix.
The high-resolution bounding box images are used to train the super-resolution CNN model.
To restore the original image, the transform $\mathbf{T}^{-1}$ is applied to the cropped image.

\begin{figure*}[t]
    \centering
    \includegraphics[width=0.99\linewidth]{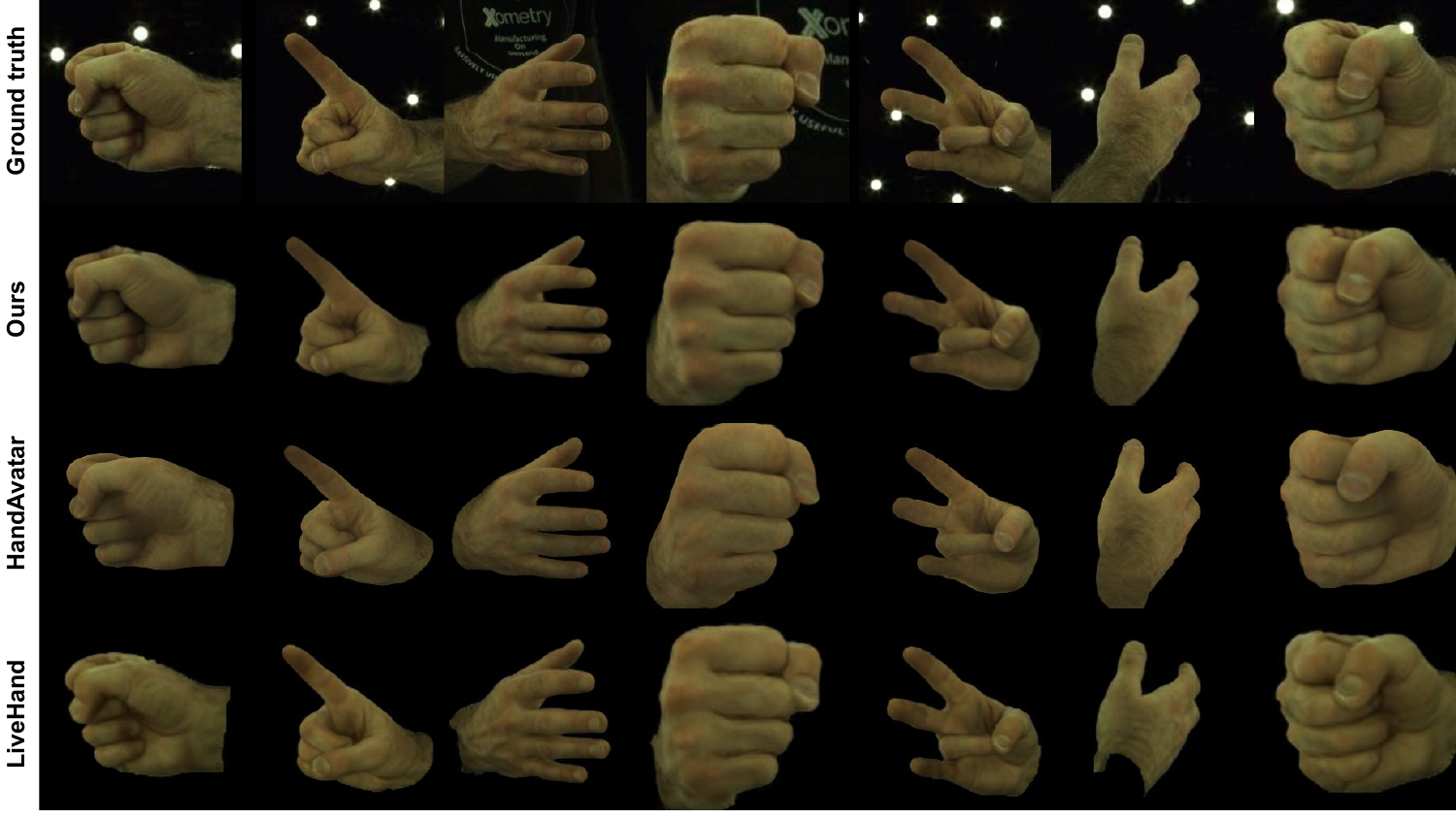}
    \caption{Comparison of single hand rendering to LiveHnad~\cite{mundra2023livehand} and HandAvatar~\cite{bib:handavatar} methods.}
    \label{fig:comparison_single_hands}
\end{figure*}
\begin{figure*}[t]
    \centering
    \includegraphics[width=0.99\linewidth]{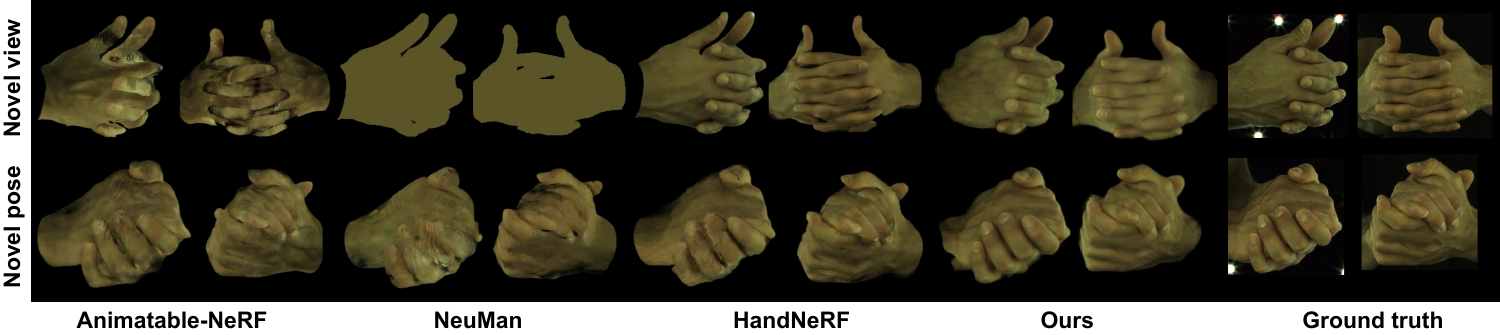}
    \caption{Comparison of interacting hands with Animatable-NeRF~\cite{peng2021animatable}, NeuMan~\cite{jiang2022neuman}, and HandNeRF~\cite{Guo_2023_CVPR}. Each pair of images shows two different views of the same pose. The first row applies NeRF to a novel view, while the second row applies it to a novel pose. The competitor images are sourced from the HandNeRF.}
    \label{fig:comparison_interacting_hands}
\end{figure*}


\begin{figure*}[t]
    \centering
    \begin{subfigure}[t]{0.428\textwidth}
        \centering
        \includegraphics[width=0.99\linewidth]{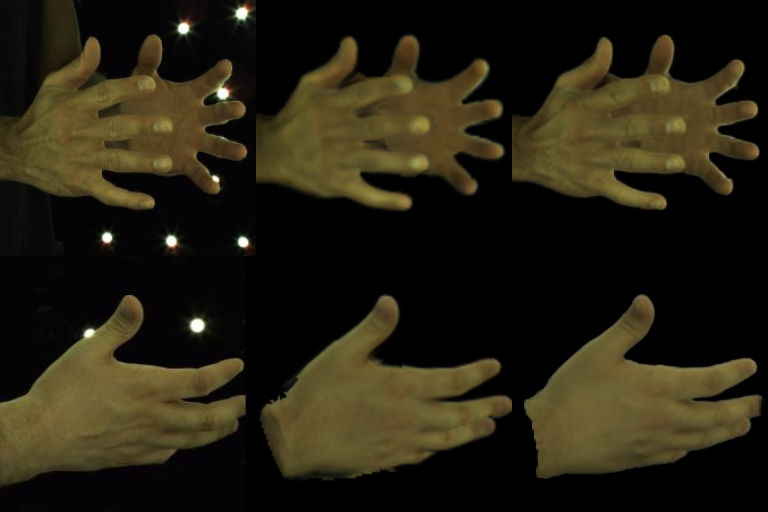}
        \caption{}
    \end{subfigure}%
    ~
    \begin{subfigure}[t]{0.572\textwidth}
        \centering
        \includegraphics[width=0.99\linewidth]{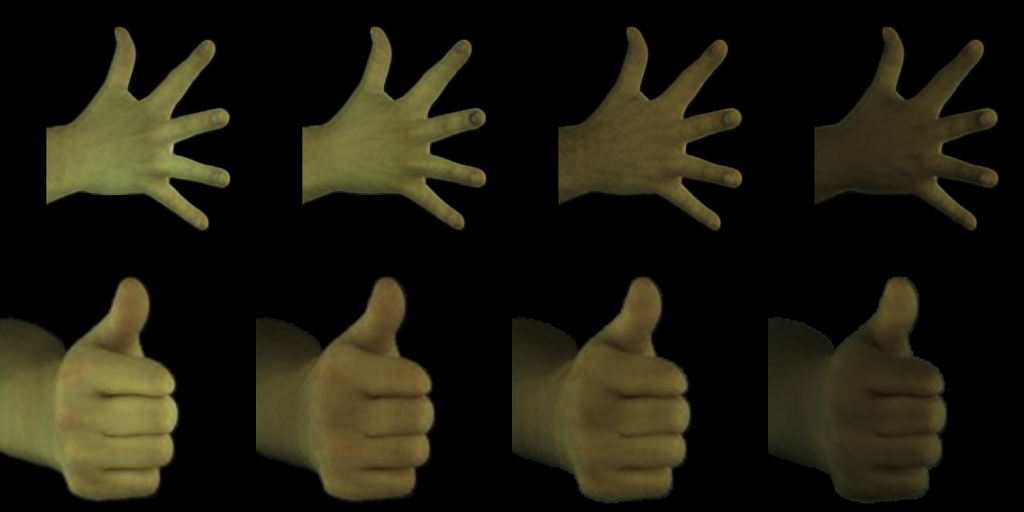}
        \caption{}
    \end{subfigure}
    \caption{Figure (a) shows: ground truth, output from the NeRF, output from the upscaling CNN. Figure (b) demonstrates appearance transfer of the same hand pose to different identities.}
    \label{fig:nerf2cnn}
\end{figure*}

\removetext{To preserve the original image coordinates, the intrinsic matrix is updated as follows $\mathbf{K}^\prime \leftarrow \mathbf{T} \mathbf{K}$.
The down-scaled (by factor $k$) bounding box images are used for NeRF training, where the focal lengths and principal point of $\mathbf{K}^\prime$ are further divided by $k$.}
\removetext{The whole pipeline is demonstrated in Fig.}

\subsection{Quantitative evaluation}


\begin{table}[t]
    \setlength{\tabcolsep}{17pt}
    \centering
    \caption{Comparison of rendering quality on~{\intdataset}. The symbol $^*$ indicates that the method was the version implemented in the LiveHand paper~\cite{mundra2023livehand}. 
    }
    \vspace{-0.1cm}

    \begin{tabular}{lrcrr}
    \hline
         &  PSNR $\uparrow$ & LPIPS (x1000)  $\downarrow$ $\uparrow$ \\ \hline
         Mesh wrapping & 28.28 & 49.44  \\ 
         SMPLpix~\cite{prokudin2021smplpix} & 32.37 & 26.57 \\ 
         A-NeRF$^*$~\cite{su2021anerf} & 28.07 & 94.41  \\ 
         LISA$^*$~\cite{corona2022lisa} & 29.36 & 78.46  \\ 
         LiveHand~\cite{mundra2023livehand} & 32.04 & {\bf 25.73} \\ 
         Ours & {\bf 32.38} &  27.92 \\ \hline
    \end{tabular}
    \label{tab:livehand_comparison}
\end{table}
\begin{table*}[t]
    \setlength{\tabcolsep}{13pt}
    \centering
    \caption{Comparison of rendering results on single and interacting hands. The columns represent models trained on 4, 7, and 10 views, respectively, and tested on 18 views.
    The symbol $^*$ denotes models implemented by HandNeRF~\cite{Guo_2023_CVPR}. The symbol $\times$ indicates that the model failed to converge.}
    \vspace{-0.1cm}

    \begin{tabular}{l|rrr|rrr|rrr}
    \hline 
    & \multicolumn{3}{c}{4 views} & \multicolumn{3}{c}{7 views} & \multicolumn{3}{c}{10 views} \\

         &  PSNR  & SSIM   & LPIPS   &  PSNR  & SSIM   & LPIPS   &  PSNR  & SSIM   & LPIPS     \\ \hline

     \multicolumn{10}{c}{Single hand} \\ \hline 

        Pose-NeRF$^*$~\cite{Barron2021MipNeRFAM} & 27.085 & 0.935 & 0.092 & 29.264 & 0.930 & 0.070 & 29.212 & 0.939 & 0.073 \\ 
        Ani-NeRF$^*$~\cite{peng2021animatable} & 30.260 & 0.958 & 0.070 & 31.642 & 0.963 & 0.058 & 31.778 & 0.968 & 0.062 \\
        NeuMan$^*$~\cite{jiang2022neuman} & 30.342 & 0.959 & 0.069  & 31.236 & 0.962 & 0.057  & 31.841 & 0.970 & 0.055 \\
        HandNeRF~\cite{Guo_2023_CVPR} & 31.049 & {\bf 0.965} & 0.058 & 31.855 & {\bf 0.969} & 0.045 & 32.703 & {\bf 0.974} & {\bf 0.037}  \\


       Ours & {\bf 31.958}  & 0.962 & {\bf 0.043} &  {\bf 32.741} & 0.965 & {\bf 0.039} & {\bf 33.090} & 0.967 &  0.038 \\ \hline

        \multicolumn{10}{c}{Interacting hands} \\ \hline 
        Pose-NeRF$^*$~\cite{Barron2021MipNeRFAM} & 25.019 & 0.874 & 0.187 & 27.241 & 0.901 & 0.138 & 27.646 & 0.916 & 0.107 \\
        Ani-NeRF$^*$~\cite{peng2021animatable} & 28.032 & 0.941 & 0.086 & 28.854 & 0.944 & 0.084  & 29.357 & 0.949 & 0.079  \\
        NeuMan$^*$~\cite{jiang2022neuman} & $\times$ & $\times$ & $\times$ & $\times$ & $\times$ & $\times$ & $\times$ & $\times$ & $\times$ \\
        HandNeRF~\cite{Guo_2023_CVPR} & 29.035 & {\bf 0.955} & 0.084 & 30.069 & {\bf 0.962} & 0.081 & 30.757 & {\bf 0.956} & 0.072 \\

       Ours & {\bf 29.746} & 0.931 & {\bf 0.079} & {\bf 30.706} & 0.938 & {\bf 0.071} & {\bf 30.836} & 0.939 & {\bf 0.070} \\ \hline

    \end{tabular}
    \label{tab:lhandnerf_comparison}
\end{table*}

\begin{table*}[t]
    \setlength{\tabcolsep}{13pt}
    \centering
    \caption{Comparison of rendering quality on the HandAvatar~\cite{bib:handavatar} splits (first row) of the~{\intdataset} dataset. A higher SSIM is better.   }
    \begin{tabular}{l|ccc|ccc|ccc}
    \hline 
    & \multicolumn{3}{c|}{\emph{test/Capture0}} & \multicolumn{3}{c|}{\emph{test/Capture1}} & \multicolumn{3}{c}{\emph{val/Capture0}} \\


        \multirow{-2}{*}{Method} & LPIPS    &  PSNR  & SSIM   & LPIPS   &  PSNR  & SSIM  & LPIPS  &  PSNR  & SSIM   \\ \hline
        

        SelfRecon~\cite{jiang2022selfrecon} & 0.142 & 26.38 & 0.878 & 0.138 & 25.18 & 0.875 & 0.149 & 25.78 & 0.868 \\
        HumanNeRF~\cite{weng_humannerf_2022_cvpr} & 0.114 & 27.64 & 0.883 & 0.117 & 26.31 & 0.880 & 0.119 & 27.80 & 0.881 \\
        HandAvatar~\cite{bib:handavatar} & {\bf 0.103} & 28.23 & 0.894 & {\bf 0.107} & 26.56 & {\bf 0.890} & {\bf 0.106} & 28.04 & 0.890\\
       Ours & 0.107 & {\bf 28.33}  & {\bf 0.895} & 0.117 & {\bf 26.62} & 0.886 &  0.112 & {\bf 28.23} & {\bf 0.891} \\ \hline
    \end{tabular}

    \label{tab:handavatar_comparison}
\end{table*}
\begin{table}[t]
    \setlength{\tabcolsep}{10pt}
    \centering
    \caption{The comparison results on the HanCo~\cite{Freihand2019,ZimmermannAB21} dataset.}
    \begin{tabular}{lccc}
    \hline
         &  PSNR $\uparrow$ & LPIPS (x1000)  $\downarrow$ & SSIM  $\uparrow$ \\\hline
         HandAvatar~\cite{bib:handavatar} & 18.712 & {\bf 16.231} & {\bf 0.844} \\ 
         Ours & {\bf 19.181} &  16.320 & 0.842  \\ \hline
    \end{tabular}
    \label{tab:hanco_comparison}
\end{table}
Table~\ref{tab:livehand_comparison} compares our method against the~{\sota} approaches.
The evaluation protocol follows the instructions outlined in the LiveHand description~\cite{mundra2023livehand}. 
We outperform other approaches except for the LPIPS metric.
This is because the ground truth images are masked with the MANO mesh, and since our method does not utilize mesh information, the LPIPS metric becomes sensitive to the overall hand shape.


We followed the evaluation guidelines in HandNeRF~\cite{Guo_2023_CVPR} and evaluated model performance trained with 4, 7, and 10 views on 18 different test views.
The results of the experiments are shown in Table~\ref{tab:lhandnerf_comparison}.
We outperform the~{\sota} except for the SSIM metric, where we are comparable.

By applying the training and evaluation protocol from the HandAvatar~\cite{bib:handavatar} method, the comparison results on different splits of the~\intdataset dataset are presented in Table~\ref{tab:handavatar_comparison}. 
On most metrics, our approach is either better or comparable to the~{\sota}.

The results on HanCo~\cite{Freihand2019,ZimmermannAB21} are shown in Table~\ref{tab:hanco_comparison}.
We used the HandAvatar code and trained the method on all images of persons 26 and 29 without hand-to-object interactions. 
The dataset is very challenging and has a lot of illumination changes, hence, the perceptual accuracy is rather low.
However, our method outperforms the HandAvatar method on the PSNR and is marginally behind on LPIPS and SSIM metrics.

\newstuff{For a fair evaluation, the main input to our rendering model is a 3D skeleton provided in the dataset, as the competitors use the ground truth MANO mesh parameters. The skeleton carries less information, specifically in describing hand volume, hence leaving our method with a big disadvantage. However, despite this, the proposed model still outperforms the competitors on many metrics, and only marginally yields to the others.
Additionally, we artificially introduced small Gaussian noise to the input 3D skeleton to simulate a network prediction; and we find that it has negligible impact on the model's accuracy.
}

\subsection{Qualitative evaluation}
Fig.~\ref{fig:nerf2cnn} illustrates the effect of the upscaling CNN on the NeRF output.
Additionally, it demonstrates appearance transfer between different identities by providing corresponding CVAE embeddings to NeRF.

Qualitative comparisons of hand images rendered by the proposed approach to the \emph{state-of-the-art} are demonstrated in Fig.~\ref{fig:comparison_single_hands} and Fig.~\ref{fig:comparison_interacting_hands}. The figures show both the rendered hand and the ground truth image for visual comparison. Here we see excellent reproduction capability. There is some loss of detail, especially around the nails, and some smoothing. But on the whole, the results are visually very close to the natural images.

\subsection{Limitations}

One of the limitations of our approach is a longer training time compared to hand parametric-based methods.
The reason for this is that the parametric model reduces the problem's dimensionality, allowing the model to converge faster.

Additionally, extracting appearance and shape from a single image is a challenging task. 
We use CVAE latent embeddings that demonstrate good generalization, however, it may not always be accurate due to the single-view ambiguity.

\section{CONCLUSIONS}

We present a novel framework for 3D hand rendering that exploits a NeRF renderer that generalizes across multiple views and hand poses.
The proposed method avoids the hard constraint of initialization and/or a parametric mesh model, widely adopted in the literature.
Instead, we provide a step-by-step pipeline starting from point cloud extraction, and training of conditioned occupancy probabilities which are then combined into a NeRF as an implicit shape model to render 3D hands.
The hand geometry is represented via occupancy probabilities and features, while appearance and shape are extracted and parametrized via a latent vector extracted from the image via a CVAE.
The proposed NeRF conditioning combines these elements to efficiently render novel poses and views. 
On the benchmark publicly available~{\intdataset} dataset, our method achieves~{\sota} accuracy.

\section{ACKNOWLEDGMENTS}
This work was supported by the SNSF project `SMILE II' (CRSII5 193686), the Innosuisse IICT Flagship (PFFS-21-47), EPSRC grant APP24554 (SignGPT-EP/Z535370/1) and and through funding from Google.org via the AI for Global Goals scheme. This work reflects only the author’s views and the funders are not responsible for any use that may be made of the information it contains.


\section*{ETHICAL IMPACT STATEMENT}
In this paper, we used two publicly available benchmark hand datasets: InterHand2.6M~\cite{Moon_2020_ECCV_InterHand2.6M} and HanCo~\cite{Freihand2019,ZimmermannAB21}. These datasets do not include any identifiable faces only images of hands, 3D keypoints, MANO mesh parameters, and segmentation masks. Additionally, the datasets are diverse in race and gender. We believe that using this data for quantitative and qualitative evaluation carries minimal risk of harm to participants originally captured in the datasets. We strictly follow the dataset protocols that use numerical identifiers to represent ground truth, maintaining privacy and respecting participant anonymity.

We do not anticipate any negative societal impacts from our research. The goal of this paper is solely to present a more efficient rendering method, which we believe will benefit research in computer vision. As with any artificial intelligence tool, there is a potential risk of the model being misused to blur the distinction between real and AI-generated hands. However, our intention is quite the opposite; by developing techniques for identity change using CVAE embeddings, our work aims to support anonymization. 
Moreover, in the provided visual figures, we demonstrate performance across different races to avoid potential bias or discrimination.
This approach could reduce the need for real data collection, substituting it with rendering techniques that preserve privacy, which may help alleviate some of the ethical concerns associated with biometric data usage.







{\small
\bibliographystyle{ieee}
\bibliography{egbib}
}
\clearpage




\section{Implementation Details}

\begin{figure}
    \centering
    \includegraphics[width=0.32\linewidth]{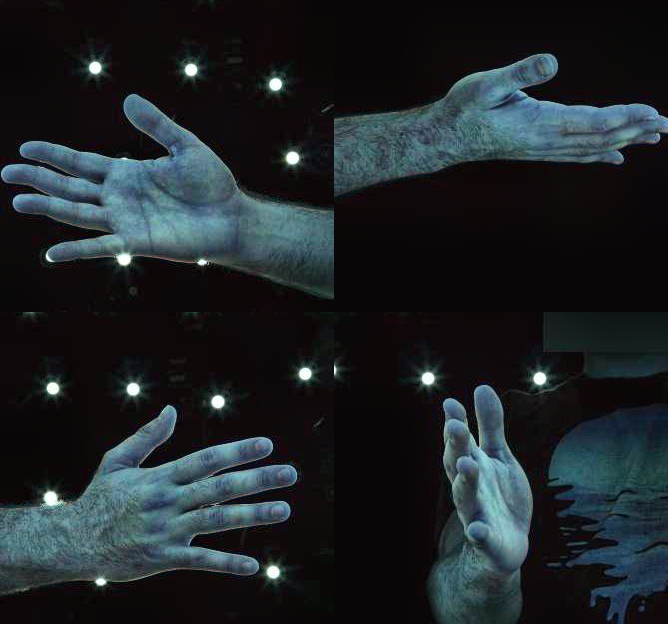}
    \includegraphics[width=0.32\linewidth]{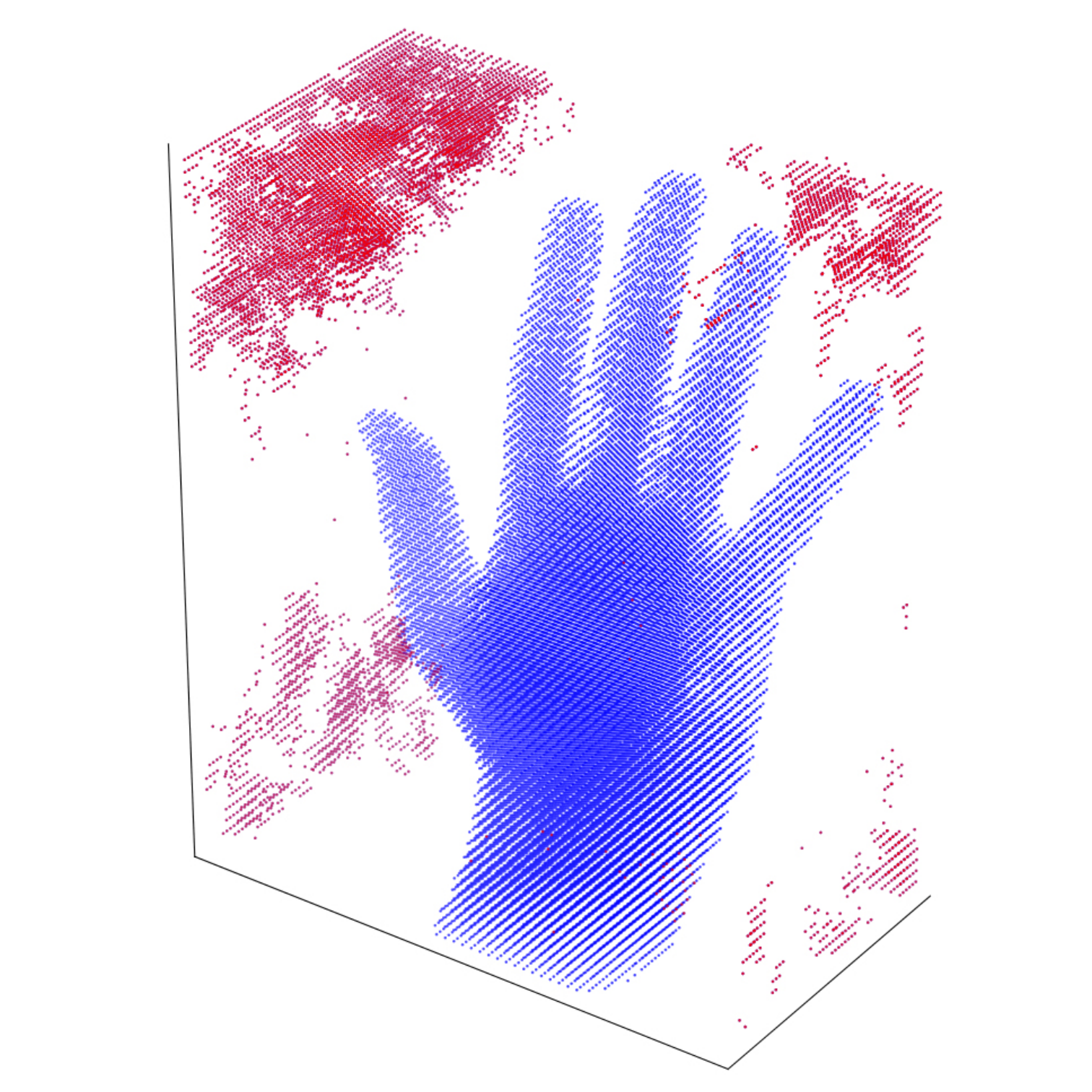}
    \includegraphics[width=0.32\linewidth]{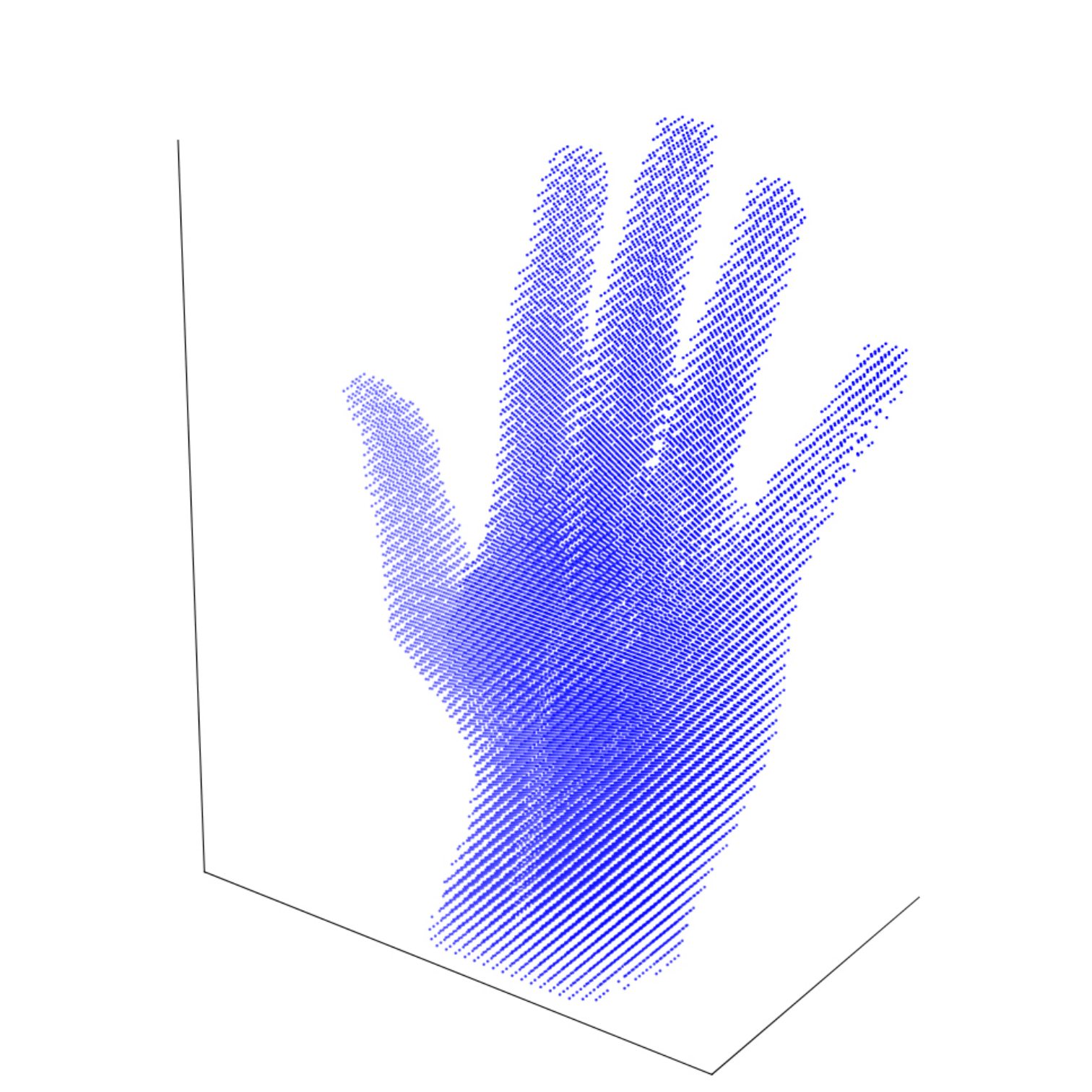}
    \caption{The left figure shows a collage of different hand images in ``normalized'' image space. The central figure shows the corresponding 3D hand point cloud extracted using the color consistency, where red points highlight outliers that randomly have the same color on all images. The last figure demonstrates a hand point cloud if all images were masked. }
    \label{fig:point_clouds}
\end{figure}

{\bf Volumetric Hand Cloud.} The point cloud extraction process is visualized in Fig.~\ref{fig:point_clouds}.
We generate a point cloud and project it onto the multi-view images.
By checking color consistency and using hand segmentation masks, we can filter out background points.

{\bf CVAE.}
Directly using the hand skeleton as input to the CVAE is impractical due to its low resolution and vector format. Instead, we generate an image representation of the hand skeleton by rendering its vertices and edges. These hand skeleton images are then concatenated with real images (resulting in a 6-channel image) and fed into the CVAE.

{\bf Occupancy Conditioning.}
For interacting hands, we condition the NeRF using the probability occupancies of both hands. If \( p_r \) and \( p_l \) represent the probabilities of a point belonging to the right and left hand, respectively, the encoded value is \( p_r \) if \( p_r > p_l \), and \(-p_l\) otherwise.

\begin{figure}
    \centering
    \includegraphics[width=0.99\linewidth]{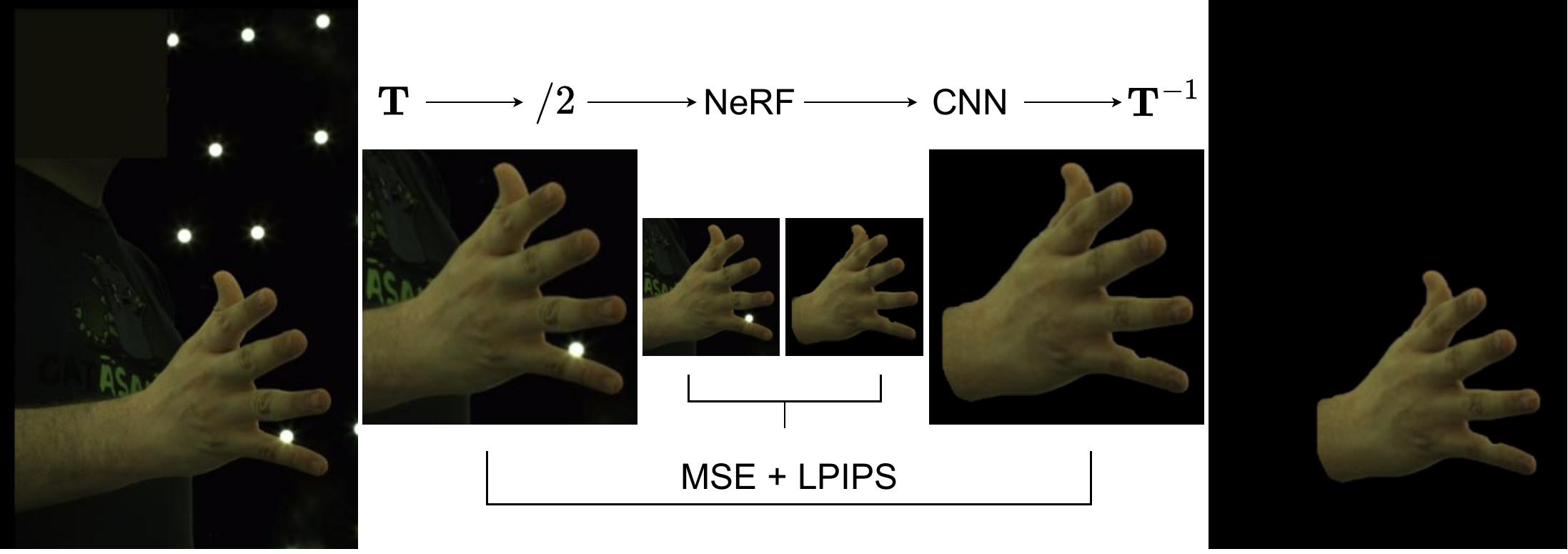}
    \caption{Pipeline demonstrating the rendering process. The original image is first warped with an affine transform $\mathbf{T}$. It is then downscaled by a factor $k=2$. The NeRF renders a downscaled image, which is then upscaled by a CNN model. Finally, using the inverse affine transform $\mathbf{T}^{-1}$, the image in the original resolution is obtained. The LPIPS and MSE loss are applied only to bounding boxes and downscaled images.}
    \label{fig:affine_transform}
\end{figure}

{\bf Image Transform.}
To preserve the original image coordinates, the intrinsic matrix is updated with the affine transform \( \mathbf{T} \) as follows: \( \mathbf{K}^\prime \leftarrow \mathbf{T} \mathbf{K} \). The down-scaled (by factor \( k \)) bounding box images are used for NeRF training, where the focal lengths and principal point of \( \mathbf{K}^\prime \) are further divided by \( k \). The entire pipeline is demonstrated in Fig.~\ref{fig:affine_transform}.

\section{Training and evaluation data}

In our experiments, we used two datasets and followed different evaluation protocols from state-of-the-art methods:

\begin{itemize}
\item 
The LiveHand~\cite{mundra2023livehand} protocol uses only right-hand images selected from the ``train/capture0'', ``train/capture5'', ``test/capture0'', and ``test/capture1'' subsets of the \textsc{InterHand2.6M} dataset. The last 50 frames of each capture were reserved for evaluation, with the rest used for training.

\item 
The HandAvatar~\cite{bib:handavatar} evaluation protocol reserves right-hand images from the \emph{ROM04\_RT\_Occlusion} sequence for training and the \\\emph{ROM03\_RT\_No\_Occlusion} sequence for evaluation, across the \\ ``test/capture0'', ``test/capture1'', and ``val/capture0'' subsets.

\item 
HandNeRF~\cite{Guo_2023_CVPR} uses single and interacting hand data from the 30FPS \textsc{InterHand2.6M} dataset, specifically from the ``train/capture0'' subset. Detailed sequences and views are provided in the supplementary material of the HandNeRF.

\item 
For the HanCo~\cite{Freihand2019,ZimmermannAB21} dataset, we used 26,636 training images and 6,956 test images, filtering out identities 26 and 29 and excluding images with object interaction.
\end{itemize}
For training, we additionally employed a perceptual LPIPS loss with a VGG network~\cite{vgg}, and for testing, we used an AlexNet~\cite{NIPS2012_c399862d} backbone.


\begin{table}
    \setlength{\tabcolsep}{13pt}
    \centering
    \caption{Comparison between the NeRF model trained with conditioning on hand appearance embeddings (first row, ``w'') and without (second row, ``w/o'').}
    \begin{tabular}{l|c|c|c}
    \hline
         &  LPIPS $\downarrow$ & PSNR $\uparrow$ & SSIM $\uparrow$ \\ \hline
         w/\phantom{o} appearance & {\bf 0.04145 } & {\bf 34.201 } & {\bf 0.9706 }\\
         w/o appearance &  0.04410  & 33.733 & 0.9703 \\ \hline         
    \end{tabular}
    \label{tab:app}
\end{table}

\begin{table*}
    \setlength{\tabcolsep}{20pt}
    \centering
    \caption{Comparison between the NeRF model trained with conditioning on occupancy probabilities (first row, ``w'') and without (second row, ``w/o'') on single and interacting hands.}
    \begin{tabular}{l|r|r|r||r|r|r}
    \hline
         & \multicolumn{3}{c||}{Single Hand} & \multicolumn{3}{c}{Interacting Hands} \\ \cline{2-7}
         &  LPIPS $\downarrow$ & PSNR $\uparrow$ & SSIM $\uparrow$ & LPIPS $\downarrow$ & PSNR $\uparrow$ & SSIM $\uparrow$ \\ \hline
         w/\phantom{o} probabilities & {\bf 0.05632} & {\bf 32.117} & {\bf 0.9581} & {\bf 0.07418} & {\bf 31.465} & {\bf 0.9426} \\
         w/o probabilities & 0.05863 & 32.076 & {\bf 0.9581} & 0.07872 & 31.354 & 0.9417 \\ \hline         
    \end{tabular}
    \label{tab:occ}
\end{table*}

\section{Ablation studies}

For the ablation studies, we carefully selected representative subsets of the \textsc{InterHand2.6M} dataset~\cite{Moon_2020_ECCV_InterHand2.6M}.
Firstly, we investigate the influence of occupancy probabilities on NeRF performance.
We selected approximately 32,616 images for training and 5,256 for evaluation. In the evaluation set, there are 3,891 instances of interacting hands and 1,365 instances of single hands.

Secondly, we explore the impact of appearance embeddings on NeRF's accuracy. 
We utilized 55,827 images for training and 5,903 for evaluation, containing all person identities from the \textsc{InterHand2.6M} dataset.
Some identities were employed for both training and evaluation, while others were exclusively reserved for evaluation.

Comparison results of conditioning the NeRF model with and without occupancy probabilities are presented in Table~\ref{tab:occ}.
The results empirically confirm that conditioning with occupancy probabilities enhances NeRF's accuracy across various metrics, with a particularly notable improvement observed for interacting hands.

Similarly, comparison results of conditioning the NeRF model with and without hand appearance embeddings are demonstrated in Table~\ref{tab:app}.
Likewise, the results for the model trained with appearance embeddings outperform those without, as the embeddings offer additional information to the network regarding target identity.

\end{document}